\documentclass[11pt]{article}

\usepackage[T1]{fontenc}
\usepackage{lmodern}
\usepackage[letterpaper,margin=1in]{geometry}
\usepackage{amsmath,amssymb,amsthm}
\usepackage{graphicx}
\usepackage{booktabs}
\usepackage{array}
\usepackage{longtable}
\usepackage{multirow}
\usepackage{xcolor}
\usepackage{enumitem}
\usepackage[colorlinks=true,linkcolor=blue!60!black,citecolor=blue!60!black,urlcolor=blue!60!black]{hyperref}
\usepackage{caption}
\AtBeginDocument{\DeclareFontShape{T1}{lmr}{bx}{sc}{<->ssub*lmr/m/sc}{}}
\newcommand{\tb}{\textsc{TruthInsightBench}}
\newcommand{\rcb}{\textsc{ResearchClawBench}}

\title{\tb: An Evidence-Grounded Benchmark for Automated Evaluation of Open-Ended Scientific Discovery Agents}

\author{%
TruthInsight-AI\thanks{Correspondence: \texttt{TruthInsightBench@163.com}.}
}

\date{}

\begin{document}
\maketitle

\begin{abstract}
Autonomous coding agents are increasingly proposed as AI-scientist systems capable of conducting analyses and authoring research reports. Yet carrying out a prescribed analysis is not the same as making a scientific discovery, and current benchmarks do not distinguish the two: they are designed for reproduction, constructing tasks, data, and rubrics around a hidden target study and rewarding recovery of its result. We present \tb, a benchmark designed for discovery. Its 40 blind tasks, derived from 40 peer-reviewed studies across 10 scientific domains, pose open-ended discovery problems: an agent receives only a neutral scientific objective and frozen data, while source conclusions, expected values, and analysis paths are withheld, leaving it to determine what claim the data support. A fixed LLM-based judge assesses the evidentiary maturity of an agent's own scientific claims along six dimensions, operationalized as 29 artifact-grounded items; aggregation is automated and deterministic, with no per-instance human grading, so that evaluation can be repeated automatically as agents evolve in self-improving research loops. Evaluating four coding agents on one frozen base model reveals a consistent pattern. Total scores form a narrow plateau (58.4--60.3 of 100) with no statistically reliable pairwise separation, at the level of execution rather than discovery. The agents conduct and report analyses competently: evidence auditability and novelty are comparatively strong, whereas the discriminating acts that establish a trustworthy claim (controls, robustness, falsifiability, and cross-dataset generalization) are largely absent. The bottleneck is thus scientific judgment rather than coding: agents carry out analytical workflows whose conclusions they cannot adjudicate, and genuine scientific discovery remains out of reach. \tb{} makes this gap a measurable target for progress as AI-scientist agents evolve; the task data and scoring program are at \texttt{https://github.com/TruthInsight-stack/TruthInsightBench}.
\end{abstract}

\section{Introduction}
\label{sec:intro}

Automated scientific research is emerging as a central frontier of artificial intelligence (AI) \cite{douglas2025}. Coding agents such as Claude Code \cite{claudecode2026}, Codex CLI \cite{codexcli2026}, and their open-source analogues \cite{openscience2026,deepseekharness2026} are increasingly marketed as autonomous research (``Auto-Research'') tools. Yet the scientific community still lacks a principled way to ask whether such claims hold under scrutiny, and, more importantly, what these claims actually imply about scientific ability.

We argue that two distinct abilities are being conflated. The first is the ability to \emph{execute a defined analysis}: given a scientific question, a dataset, and an expected form of output, to write code, run the prescribed analyses, produce figures, and report the expected result. This is an engineering capability, necessary but not sufficient for science, and it is what most existing benchmarks measure. The second is the capacity for \emph{open-ended scientific inquiry}: to examine raw data without a known answer, determine which phenomena merit a claim, separate a genuine effect from a numerical or modeling artifact, formulate falsifiable hypotheses, adjudicate between competing explanations, test controls and robustness, establish that a result holds on independent data, delimit the scope within which a conclusion holds, and revise that conclusion when counter-evidence arrives. This second capacity is what distinguishes discovery from execution across scientific domains and data modalities, and it is largely absent from current evaluation.

A second requirement follows from the objective of self-improving research agents, namely systems that conduct and iteratively refine their own research, a loop sometimes termed \emph{recursive self-improvement} (RSI)~\cite{good1965}. Such a loop can be sustained only if its feedback signal is itself automatic: a discovery score that requires human experts to grade every new instance cannot scale and therefore cannot support repeated self-improvement. Its scoring must therefore be \emph{automated and artifact-grounded}: the evidence for each item must be located in the agent's actually executed analyses and produced artifacts, and a frozen judge applies the rubric to that evidence rather than scoring free-form similarity; only score aggregation is fully deterministic. Human expertise is then used to calibrate and audit this automated signal, not to bottleneck every evaluation.

The gap is one of benchmark \emph{configuration}, that is, how the task setting, evaluation object, and scoring are arranged, rather than of difficulty, and reproduction remains a legitimate design in its own right. Across reproduction, end-to-end research, and open-ended analysis, most existing scientific-agent benchmarks build the task, data, and rubric around a hidden target study: the task situates the agent in a known phenomenon, the data are those of that study, and the rubric decomposes its artifacts, or expert-authored ground-truth decisions, into criteria the output is expected to recover \cite{paperbench2025,corebench2024,scienceagentbench2025,blade2024,firebench2026}. This is well suited to faithful reproduction and target-anchored re-discovery, but it is less naturally matched to open-ended discovery, for two reasons. The endpoint is fixed in advance, leaving little room to credit a well-supported claim the source study never made. More relevant to automation, assembling each such dataset typically calls for substantial expert effort: its reference rubric, expected quantities, and ground-truth decisions are hand-authored and verified for every source paper, and outputs are then assessed against that task-specific key. When the required human effort grows with every dataset and every graded instance, evaluation does not readily run unattended or at scale, which is in tension with automated, iteratively improvable research.

We therefore explore the complementary setting: rather than merely re-tuning the passing threshold of a reproduction-oriented benchmark, we change all three components (the task setting, the evaluation object, and the scoring) to address discovery directly, while keeping scoring automatic so that admitting a new task does not entail authoring a bespoke answer key. We offer this as an exploratory first attempt, complementary to, rather than a replacement for, reproduction benchmarks, which answer a different and necessary question.

We present \tb, a discovery-oriented benchmark built on this principle. \tb\ contains 40 blind tasks derived from 40 peer-reviewed studies published between 2018 and 2024 and spanning 10 scientific domains. Each task provides a system with a neutral research goal and a frozen set of raw data; the source study is used only during construction to verify that the data contain discoverable structure. Source conclusions, expected values, analysis paths, verification anchors, and all evaluation assets are hidden during the agent's run. The evaluation object is not ``how close is the output to a reference paper'' but ``how much evidentiary maturity does the agent's own scientific claim exhibit.'' This maturity is decomposed into six dimensions (evidence auditability, robustness, control testing, cross-dataset generalization, novelty, and falsifiability) and operationalized into 29 predefined, verifiable rubric items. Because no endpoint is privileged, null results, counterexamples, method artifacts, and scope boundaries are scored on the same footing as positive regularities, provided they are backed by the same evidentiary standards.

We use \tb\ to run a first controlled baseline study. To separate the effect of the agent \emph{scaffold} from the effect of the underlying large language model (LLM), we evaluate four mainstream autonomous coding agents (Claude Code (v2.1.220), Codex CLI (v0.149.0), OpenScience (v2.0.1)~\cite{openscience2026}, and DeepSeek Harness (v0.1.0rc7)~\cite{deepseekharness2026}), all using the same frozen base model on the same 40 tasks under the same protocol. The results reveal a consistent and, we argue, informative pattern:

\begin{itemize}[leftmargin=1.5em,itemsep=1pt]
    \item All four systems fall within a narrow interval of 58.4 to 60.3 (of 100), and no system is reliably superior to another at the task level. Under our operational rubric, all remain substantially below the standard for credible discovery set out in Section~\ref{sec:benchmark}.
    \item The four systems produce a comparable number of discoveries ($\approx$3.7--4.3 semantically de-duplicated claims per task), yet their two highest-quality discoveries attain only $\approx$44 of a possible 80 points in evidence quality. The systems thus exhibit breadth without corresponding evidence depth.
    \item The shortfall is concentrated in the acts that distinguish discovery from merely running an analysis. Systems attain most of the available points where execution suffices (roughly 35--36 of the 45 evidence-auditability points and 8--9 of the 10 novelty points), but very few where scientific discrimination is required: about 2 of 15 for control testing, 1--2 of 15 for robustness, 1--2 of 5 for falsifiability, and close to 0 of 10 for cross-dataset generalization. This shortfall is shared by all four systems rather than driven by any single one.
\end{itemize}

Our contributions are as follows:
\begin{itemize}[leftmargin=1.5em,itemsep=1pt]
    \item \textbf{A discovery-oriented benchmark.} \tb, with 40 blind tasks across 10 scientific domains, whose task setting, evaluation object, and scoring are all designed for discovery rather than reproduction, and whose construction and information-boundary protocol make this design auditable.
    \item \textbf{A verifiable discovery-quality framework.} A six-dimension, 29-item rubric framework that evaluates the evidentiary maturity of an agent's self-formed claims, including null results, counterexamples, and scope boundaries, without reference to a known endpoint, grounded in actually executed analyses and their artifacts. Because it calls for neither per-instance human grading nor a task-specific answer key, it supplies an automatable feedback signal suited to repeated agent-improvement loops.
    \item \textbf{A preliminary controlled same-model study.} Holding one frozen base model fixed across four mainstream coding agents, we use \tb{} to obtain a first empirical profile of discovery ability: scores form a narrow, statistically indistinguishable plateau in which execution-oriented dimensions remain strong while the discriminating acts of science stay weak, localizing the bottleneck in scientific judgment rather than coding; multi-seed and multi-model studies are deferred to future work.
\end{itemize}

\section{Related Work}
\label{sec:related}

Three lines of work bear on evaluating scientific agents, usefully ordered by how much of the research process they engage: static scientific knowledge and reasoning; research-process and paper-reproduction benchmarks; and autonomous research systems together with the interactive environments they act in. Each successive line moves closer to end-to-end research, yet each stops short of the capability we target, namely the evidentiary maturity of a claim an agent forms for itself from data, with no privileged answer. We situate \tb{} within this progression to make precise the gap it addresses.

\subsection{Static scientific knowledge and reasoning}
\label{sec:rel-static}
Existing evaluations begin with scientific question answering and expert-level reasoning: SciQ \cite{welbl2017}, GPQA \cite{rein2023}, MMLU-Pro \cite{mmlupro2024}, and Humanity's Last Exam \cite{phan2025}, together with university-level sets such as SciBench \cite{scibench2023}, the multidisciplinary ATLAS \cite{atlas2025}, and domain-specific suites spanning physics, chemistry, and Earth science \cite{anjum2025}. These instruments probe stored knowledge and closed-book, single-step reasoning; they neither supply raw data nor require selecting a question, running analyses, forming and testing hypotheses, and synthesizing a report. Strong performance here establishes scientific literacy rather than research conduct, so a distinct, process-level evaluation is needed.

\subsection{Research-process and paper-reproduction benchmarks}
\label{sec:rel-process}
A second line places agents in research-like workflows. SciCode \cite{scicode2024} tests scientific code generation; MLAgentBench \cite{mlagentbench2023}, MLE-bench \cite{mlebench2025}, and MLGym \cite{mlgym2025} embed agents in machine-learning experimentation, Kaggle-style engineering, and gym-style research loops. Paper reproduction is the closest analogue: PaperBench \cite{paperbench2025} has agents implement and reproduce a target paper and scores alignment with it through hierarchical rubrics; CORE-Bench \cite{corebench2024} probes computational reproducibility from supplied code and data; and the ReproduceBench suite within AutoReproduce \cite{reproducebench2025} assesses automatic reproduction of executable experiment code. ScienceAgentBench \cite{scienceagentbench2025} extracts data-driven tasks from peer-reviewed papers, and DiscoveryBench \cite{discoverybench2024} formalizes multi-step discovery over 264 tasks in six domains with a faceted rubric; in both, however, tasks remain organized around a known target or hypothesis.

A recent wave pushes toward open-ended, end-to-end research: BLADE \cite{blade2024} scores analysis decisions against expert analyses, HeurekaBench \cite{heureka2026} builds exploratory questions on published studies and repositories, FIRE-Bench \cite{firebench2026} grades rediscovery by claim-level precision and recall against the source study, AstaBench \cite{astabench2025} aggregates problems across the research process, and ScienceBoard \cite{scienceboard2025} evaluates multimodal agents on graphical- and command-line-interface (GUI/CLI) workflows, yet grading still references predefined task success, an expert analysis, or the source paper's findings. Large-scale efforts extend this line: \rcb{} \cite{rcbench} assembles end-to-end research tasks drawn from published papers across ten domains and scores agent trajectories against source-derived targets, illustrating both the scale now attainable and the continued centrality of a source-anchored target. Two limitations recur across this family and jointly motivate our design. \emph{First}, an answer key, reference analysis, or target finding is fixed in advance, so a well-supported claim the source never made is difficult to credit: evaluation measures recovery rather than discovery. \emph{Second}, that key is hand-authored and verified for every source, so human effort grows with each dataset and each graded instance, which does not scale for agents that are themselves iterating. \tb{} is complementary: it retains real papers, raw data, and end-to-end reports while removing the privileged endpoint, and makes the evidentiary maturity of agent-selected claims, rather than similarity to a reference, the object of an automated score.

\subsection{Autonomous research systems and interactive environments}
\label{sec:rel-systems}
A third line situates agents in environments they act within rather than artifacts they reproduce. ScienceWorld \cite{scienceworld2022} and DiscoveryWorld \cite{discoveryworld2024} cast science as interaction in simulated worlds where agents act, observe, hypothesize, and experiment, but abstract away real datasets and real evidentiary standards; SGI-Bench \cite{sgibench2025} probes scientist-aligned workflows, while AIRS-Bench \cite{airsbench2026} and MLR-Bench \cite{mlrbench2026} target open-ended research and the full research lifecycle. In parallel, system-level efforts (The AI Scientist \cite{aiscientist2024}, AI Co-Scientist \cite{aiscientist2025}, AI-Researcher \cite{airesearcher2025}, and InternAgent-1.5 \cite{internagent2026}) demonstrate increasingly autonomous research pipelines and sharpen the evaluation problem: as such systems proliferate, comparing them on real scientific data requires a discovery-oriented, artifact-grounded instrument that reruns without per-instance human grading. \tb{} is designed as that instrument.

\begin{table}[t]
\centering
\small
\setlength{\tabcolsep}{4pt}
\caption{\textbf{How representative benchmarks relate to open-ended scientific discovery.} We record whether a benchmark is grounded in real scientific data (Raw data), requires an end-to-end research process and report (Process \& report), imposes no privileged target answer (No fixed endpoint), scores the evidentiary maturity of an agent's own claims rather than match to a reference (Self-claim evidence), and yields a score without a task-specific human-authored key (Auto grading). A $\checkmark$ is yes, $\triangle$ partial support, and $\times$ no; Dom.\ is the number of scientific domains, counted as broad disciplinary fields rather than task themes or ML subareas. Rows are grouped by the three families discussed in this section; the axes describe design orientation rather than rank quality.}
\label{tab:compare}
\begin{tabular}{@{}>{\raggedright\arraybackslash}p{4.15cm}cccccc@{}}
\toprule
Benchmark & \shortstack{Raw\\data} & \shortstack{Process \&\\report} & \shortstack{No fixed\\endpoint} & \shortstack{Self-claim\\evidence} & \shortstack{Auto\\grading} & Dom.\\
\midrule
\multicolumn{7}{@{}l}{\emph{Static knowledge and reasoning (Section~\ref{sec:rel-static})}}\\
SciQ / GPQA / HLE / SciBench / ATLAS & $\times$ & $\times$ & $\times$ & $\times$ & $\checkmark$ & many \\
\midrule
\multicolumn{7}{@{}l}{\emph{Research-process and paper-reproduction (Section~\ref{sec:rel-process})}}\\
SciCode & $\triangle$ & $\triangle$ & $\times$ & $\times$ & $\triangle$ & 6 \\
MLAgentBench / MLE-bench / MLGym & $\checkmark$ & $\triangle$ & $\times$ & $\times$ & $\checkmark$ & 1 \\
PaperBench / CORE-Bench / ReproduceBench & $\checkmark$ & $\checkmark$ & $\times$ & $\triangle$ & $\triangle$ & 1/3/1 \\
ScienceAgentBench & $\checkmark$ & $\checkmark$ & $\times$ & $\times$ & $\triangle$ & 4 \\
DiscoveryBench & $\checkmark$ & $\checkmark$ & $\times$ & $\triangle$ & $\triangle$ & 6 \\
BLADE / Heureka / FIRE / Asta / ScienceBoard & $\checkmark$ & $\checkmark$ & $\triangle$ & $\triangle$ & $\triangle$ & many \\
\rcb & $\checkmark$ & $\checkmark$ & $\triangle$ & $\triangle$ & $\triangle$ & 10 \\
\midrule
\multicolumn{7}{@{}l}{\emph{Autonomous systems and interactive environments (Section~\ref{sec:rel-systems})}}\\
ScienceWorld / DiscoveryWorld & $\times$ & $\checkmark$ & $\triangle$ & $\times$ & $\checkmark$ & 3/5 \\
SGI / AIRS / MLR-Bench & $\triangle$ & $\checkmark$ & $\triangle$ & $\times$ & $\triangle$ & many \\
\midrule
\textbf{\tb{} (ours)} & $\checkmark$ & $\checkmark$ & $\checkmark$ & $\checkmark$ & $\checkmark$ & \textbf{10} \\
\bottomrule
\end{tabular}
\end{table}

As Table~\ref{tab:compare} shows, these axes characterize design orientation rather than rank quality: they summarize how each family is positioned relative to open-ended discovery, not how strongly it performs.

\section{\tb}
\label{sec:benchmark}

Scientific discovery, as we operationalize it, starts from observational, experimental, or computational data and has no predefined endpoint: the researcher decides which phenomena merit a claim, forms and compares candidate explanations through baselines, controls, perturbations, and alternative methods, and delimits the scope in which the conclusion holds; null effects, counterexamples, method artifacts, and applicability boundaries are all valid products. We accordingly define a scientific discovery as \emph{a scientific claim formed from task data that can be independently tested and that adds a judgment about the relationships, mechanisms, quantitative characteristics, or applicability boundaries of the studied objects}.

The quality of such a discovery rests on the relation between claim and evidence rather than on whether the claim matches a reference, and decomposes into six properties: the evidence is auditable and independently verifiable; the conclusion is stable under reasonable changes in data and analysis; competing explanations are tested through controls; the conclusion receives support across datasets; the claim is a knowledge increment over prior work; and future observations can maintain, narrow, or overturn it~\cite{national2019,casadevall2016}. These six properties become the scored dimensions introduced later in this section. The remainder of the section first formalizes a task and its information boundary, then describes source-study admission and the task-construction pipeline together with the domain and modality coverage, and finally specifies the six-dimension rubric and its aggregation into task and system scores.

\subsection{Task representation and information boundary}
\begin{figure}[t]
\centering
\includegraphics[width=0.92\textwidth]{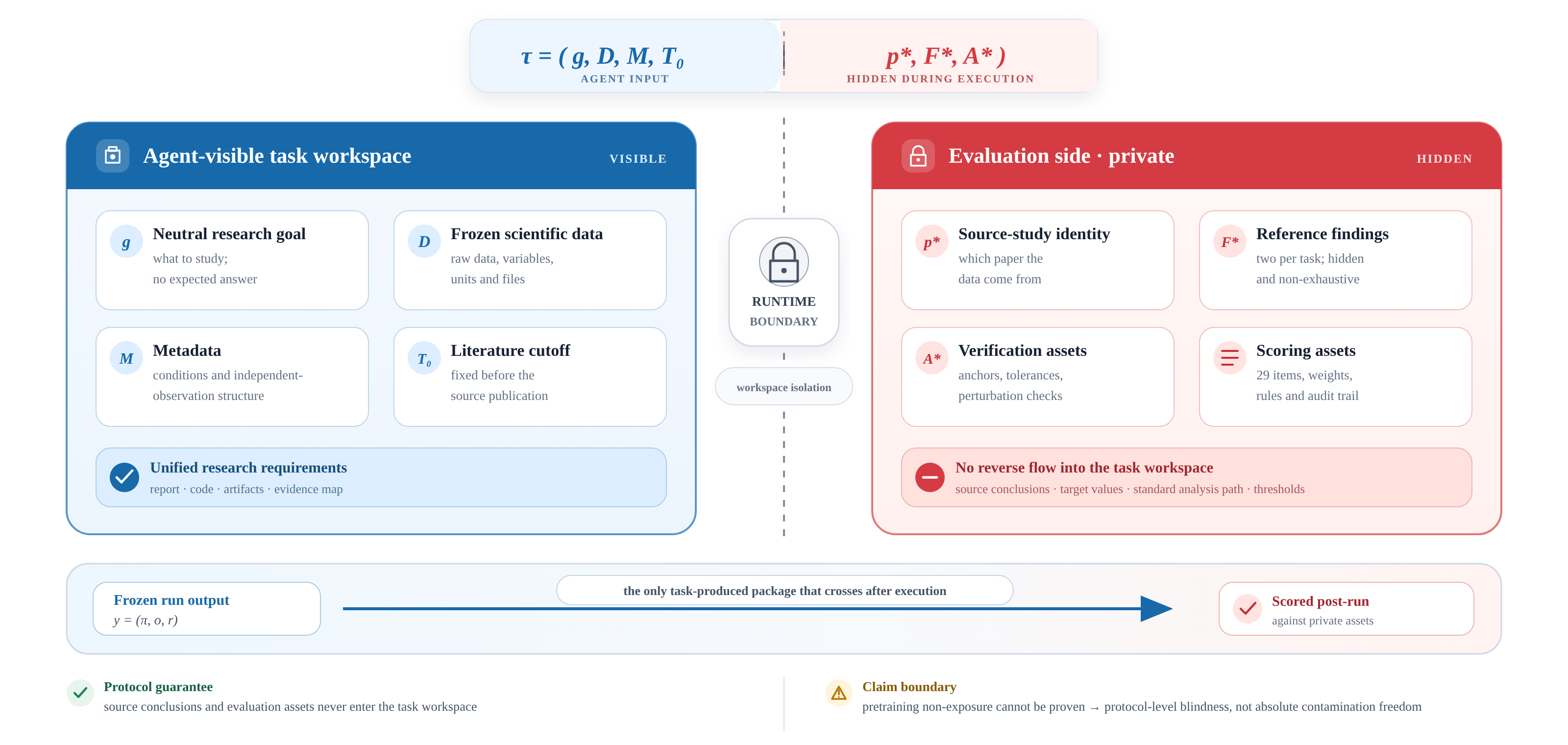}
\caption{\textbf{Information boundary of \tb.} A task $\tau=(g,D,M,T_0\mid p^{*},F^{*},A^{*})$ exposes only the agent-visible inputs (left): the neutral research goal $g$, frozen scientific data $D$, metadata $M$, the literature cutoff $T_0$, and unified research requirements. The evaluation-side assets on the right, namely the source-study identity $p^{*}$, the non-exhaustive reference findings $F^{*}$, the verification assets $A^{*}$, and the scoring items and weights, remain private during execution. The runtime boundary is one-way: only the frozen run output $y=(\pi,o,r)$ crosses after execution to be scored post-run against the private assets, while source conclusions and evaluation assets never flow back into the task workspace. We cannot guarantee that a base model never encountered related papers during pretraining, so we claim \emph{protocol-level} blindness, not absolute contamination freedom.}
\label{fig:boundary}
\end{figure}

A task is denoted
\begin{equation}
\tau \;=\; (\, g,\; D,\; M,\; T_0 \;|\; p^*, F^*, A^*\,),
\end{equation}
where $g$ is the neutral research goal, $D$ is the frozen scientific data, $M$ is the metadata (variables, units, experimental conditions, independent-observation structure), and $T_0$ is the fixed literature cutoff, set per task strictly before the source study's publication date so that its findings necessarily post-date the frozen literature window. The terms to the right of the bar are hidden: $p^*$ is the source study, $F^*$ is the non-exhaustive set of reference findings (two per task), and $A^*$ is the set of evaluation assets (verification anchors, tolerances, weights, per-item decision rules, and the audit trail), where the verification anchors are deterministic re-derivation scripts and tolerances used to recompute key quantities independently of the judge. Given task $\tau$ and an executable environment, the system produces a report $y=(\pi, o, r)$ where $\pi$ is the executed analysis code and process, $o$ are the intermediate results and artifacts, and $r$ is the final research report containing its claimed discoveries. Figure~\ref{fig:boundary} shows the information boundary.

\subsection{Source studies and task admission}
\tb{} is built from 40 peer-reviewed studies published between 2018 and 2024: 34 from \emph{Nature Communications}, 2 from \emph{Communications Physics}, 1 from \emph{Nature Physics}, 1 from \emph{npj Quantum Information}, and 2 from \emph{Science Advances}. Each study yields one task, and the 40 tasks cover Astronomy, Chemistry, Computational Mathematics, Earth Science, Energy, Information Science, Life Science, Materials Science, Neuroscience, and Physics.

Source studies are selected by \emph{purposeful curation} against three scientific admission conditions, none of which depends on the scores of any evaluated system:
\begin{enumerate}[leftmargin=1.5em,itemsep=1pt]
    \item \textbf{Interpretable data structure.} The data must retain the variable semantics, units, experimental conditions, and independent-observation structure needed to form interpretable scientific claims.
    \item \textbf{Verified recoverability.} Using only the data that will be given to the system, the builders must be able to re-derive at least two main scientific results, confirming that the data contain a discoverable scientific structure.
    \item \textbf{Testable space.} The data must contain genuine repeats, condition variation, controls, or alternative analyses, so that systems can probe uncertainty, competing explanations, and applicability boundaries.
\end{enumerate}

Solvability verification comprises, per task, one re-derivation of each of the two main results and six perturbation analyses (a sample, method, and measurement-definition change for each result), i.e. eight construction-time verifications per task and 320 across the 40 tasks. A result that cannot be re-derived from the task data alone is not used as a reference finding, and a study lacking the controls or variation structure required by the admission conditions is excluded from the official task set.

\subsection{Task construction pipeline}
Task construction proceeds in six audited steps: (1) select a real research scenario whose data support an explicit scientific question; (2) verify recoverability by re-deriving the main results from the released data alone; and (3) verify that the data contain the repeats, controls, condition variation, and alternative analyses needed for robustness and competing-explanation checks; (4) form the blind task by keeping data semantics and the neutral goal while removing the source conclusions, expected values, and standard analysis path; (5) conduct a scientific quality review covering data interpretation, main-result re-derivation, perturbation analysis, reference findings, and information isolation; and (6) freeze both the agent-visible task materials and the evaluation-side references so that all systems receive identical tasks.

The blinding covers the paper identity, conclusion direction, target values, the authors' chosen analysis route, and evaluation thresholds, while preserving what is needed to understand the data and the research scope. A scientific review confirms that blinding did not destroy variable semantics, experimental conditions, or independent-observation structure.

\subsection{Domains and data modalities}
\tb{} contains 40 tasks across 10 scientific domains (Table~\ref{tab:domains}). Task data span tabular and time-series data, structured arrays, networks, molecular or particle coordinates, and scientific images, with analysis units ranging over individuals, experimental repeats, time windows, network instances, parameter combinations, and spatial units. This heterogeneity forces systems to choose analysis methods according to the scientific object at hand rather than relying on a single statistical template.

\begin{table}[t]
\centering
\small
\caption{\textbf{Domain composition of \tb.} 40 tasks across 10 scientific domains.}
\label{tab:domains}
\begin{tabular}{lcc}
\toprule
Domain & Tasks & Ratio \\
\midrule
Astronomy & 3 & 7.5\% \\
Chemistry & 3 & 7.5\% \\
Computational Mathematics & 4 & 10.0\% \\
Earth Science & 4 & 10.0\% \\
Energy & 4 & 10.0\% \\
Information Science & 4 & 10.0\% \\
Life Science & 5 & 12.5\% \\
Materials Science & 4 & 10.0\% \\
Neuroscience & 4 & 10.0\% \\
Physics & 5 & 12.5\% \\
\midrule
Total & 40 & 100\% \\
\bottomrule
\end{tabular}
\end{table}

\subsection{Six discovery-quality dimensions}
We evaluate a single discovery along six dimensions (Table~\ref{tab:dimensions}): evidence auditability (45 points), robustness (15 points), control testing (15 points), cross-dataset generalization (10 points), novelty (10 points), and falsifiability (5 points), which sum to 100. The six dimensions organize into three blocks: current-evidence quality (75 points: evidence auditability, robustness, and control testing), confirmation and knowledge contribution (20 points: cross-dataset generalization and novelty), and future empirical testing (5 points: falsifiability). Evidence auditability carries the largest weight because a discovery must first be directly supported by current data, although evidence alone is deliberately insufficient for credible discovery. We use the term \emph{credible discovery} throughout to denote a discovery whose current-evidence dimensions (evidence auditability, robustness, and control testing) are substantially met and whose claim is both externally confirmable and falsifiable; this is an operational descriptor rather than a hard pass/fail cutoff. Robustness and control testing (15 each) evaluate stability and explanatory specificity, respectively. Cross-dataset generalization and novelty (10 each) evaluate external applicability (independent external validation is a standard guard on generalizability and reproducibility~\cite{rosenblatt2026}) and knowledge contribution. Falsifiability (5 points) evaluates whether the discovery supports a concrete, executable next empirical test.

\begin{table}[t]
\centering
\small
\caption{\textbf{Six discovery-quality dimensions for a single discovery.} Each dimension measures a distinct scientific property; points and design rationale are shown.}
\label{tab:dimensions}
\begin{tabular}{lp{8.2cm}c}
\toprule
Dimension & What is measured & Points \\
\midrule
Evidence Auditability & Whether data source, object, analysis process, key results, and final claim form a traceable, verifiable, and mutually consistent evidence chain. & 45 \\
Robustness & Whether the conclusion survives random reruns, sample changes, and method or threshold changes, and whether its scope of applicability is stated. & 15 \\
Control Testing & Whether simple baselines, no-signal controls, competing explanations, and confound/artifact checks establish that the explanation is specific. & 15 \\
Cross-Dataset Generalization & Whether the same claim continues to hold on independent data, time intervals, or objects, under pre-specified analysis and decision criteria. & 10 \\
Novelty & Whether the claim adds a new object, relation, quantitative result, or applicability boundary relative to pre-$T_0$ literature. & 10 \\
Falsifiability & Whether the claim specifies a subsequent observation or experiment that can distinguish it from competing hypotheses, and what would maintain, narrow, or overturn it. & 5 \\
\bottomrule
\end{tabular}
\end{table}

\subsection{Aggregating discoveries into task and system scores}
\label{sec:aggregation}
Across the six dimensions there are 29 verifiable items in total (8, 5, 5, 4, 2, and 5 for evidence auditability, robustness, control testing, cross-dataset generalization, novelty, and falsifiability, respectively; Appendix~\ref{app:items}). Each item is judged on three levels: \emph{fully satisfied} (completion $1$), \emph{partially satisfied} ($0.5$), or \emph{unsatisfied} ($0$). For a discovery $d$, its quality score is
\begin{equation}
Q(d) \;=\; \sum_{j=1}^{29} w_j\, c_j, \qquad c_j\in\{0,0.5,1\},\quad \sum_{j=1}^{29} w_j = 100,
\end{equation}
where $w_j$ is the weight of item $j$. Unless otherwise specified, all items are scored from analyses that were actually completed: proposing that ``a test should be run'' does not count as running it; scoring requires the corresponding data, analysis, and results to be locatable and verifiable. Scoring is automated and artifact-grounded: a single frozen LLM judge applies each item to the analyses the agent actually executed and to their result artifacts, and only the subsequent aggregation is deterministic, so an evaluation reruns on a new agent version with no per-instance human grading.

Because a system may report several discoveries for the same task, the frozen judge scores every reported claim on the 29 items and then performs semantic de-duplication \emph{within each system--task report} (never across different systems, so no system is penalized or rewarded for agreeing with another), grouping claims that share the same object, scope, relation type, direction, and key evidence under a written protocol. The highest-quality group yields $S_1$ and the second, semantically distinct group yields $S_2$ (if fewer than two groups exist, $S_2=0$). The system--task score is
\begin{equation}
R_{a,t} \;=\; \mathrm{clamp}\bigl(0.60\,S_1 + 0.20\,S_2 + B_{\mathrm{yield}} + B_{\mathrm{target}} - P_{\mathrm{false}},\; 0,\; 100\bigr),
\end{equation}
where:
\begin{itemize}[leftmargin=1.5em,itemsep=1pt]
    \item The evidence-quality main body (up to 80 points) weights the top two semantically distinct discoveries at $0.60$ and $0.20$.
    \item The multi-discovery yield bonus $B_{\mathrm{yield}}$ (up to 10 points) awards $4,2,2,2$ points to the first four semantically de-duplicated discoveries whose quality reaches at least 40 of 100, a floor that requires a discovery to clear a minimal evidentiary bar before breadth counts, encouraging breadth without diluting the evidence-quality main body.
    \item The reference-coverage bonus $B_{\mathrm{target}}$ (up to 10 points) compares each of the two hidden reference findings on object, scope, relation direction, and key quantities (full coverage 5, partial 2.5, none 0). It supplements measurement of coverage but does not constitute the answer key.
    \item The deterministic-error penalty $P_{\mathrm{false}}$ (up to 10 points) deducts 5 points per semantically de-duplicated claim whose direction is clearly contradicted by actual execution or re-derivation but is still asserted in the final report.
\end{itemize}
The $0.60/0.20$ split encodes a quality-first principle. The single strongest, best-evidenced claim receives $0.60$, because a discovery benchmark should be driven primarily by how thoroughly its headline claim is established; a second, semantically distinct discovery receives $0.20$, one third as much, to credit the ability to establish an independent additional finding without letting a larger count of shallowly supported claims substitute for one deeply evidenced one. Beyond the two leading discoveries, points can enter only through the bounded bonuses (at most 0.20 of the task score combined, half of it rewarding discovery count through $B_{\mathrm{yield}}$), whereas the two leading discoveries carry 0.80; breadth is thus capped at a quarter of the evidence-quality weight. These coefficients were fixed before evaluation in the frozen protocol, and a main-body-only ablation that removes both bonuses reproduces the same system ordering, so the reported plateau does not hinge on this particular choice.
The two bonuses $B_{\mathrm{yield}}$ and $B_{\mathrm{target}}$ (at most 20 points combined) reference the source study only \emph{after} the run, during evaluation; neither the reference findings nor their decision criteria appear in the task prompt, the frozen data, or the tool environment, so they cannot orient the system toward a privileged endpoint during execution. The 80-point evidence-quality main body is instead scored from the agent's own artifacts and public pre-$T_0$ literature and does not credit recovery of the source study's conclusions. The main-body-only ablation is computable from the per-discovery component scores reported by the released scoring program.

The system score over the full benchmark is the equal-weighted mean of task scores:
\begin{equation}
B_a \;=\; \frac{1}{40}\sum_{t=1}^{40} R_{a,t}.
\end{equation}
Because 80\% of the task score is assigned to the evidence quality of the agent's own top-two discoveries, and because no endpoint is privileged, the ranking measures which system can form more reliable, more robust, more sharply delimited scientific claims from the same raw data, not which system comes closest to a known answer.

Reference convergence and discovery credibility are deliberately kept distinct. Recovering a hidden reference finding is \emph{neither necessary nor sufficient} for a high score: a novel, well-supported claim scores on the same footing as one that matches the source, whereas a claim that matches the source conclusion but lacks a traceable, robust, and controlled evidence chain remains an \emph{ungrounded discovery} and scores accordingly. $B_{\mathrm{target}}$ is small and non-gating for exactly this reason (weighting answer convergence heavily would re-introduce the answer-key logic of the reproduction configuration), and process alone cannot inflate the score, because the 45-point evidence-auditability dimension requires the executed results to actually support the asserted claim. Thus \tb{} judges how well a claim is grounded rather than whether it matches a known result: reaching the right conclusion without building its supporting case is, by design, not credited as a credible discovery.

\section{Experiments}
\label{sec:experiments}

\subsection{Experimental setup}
\label{sec:setup}

\paragraph{Systems.}
We evaluate four mainstream autonomous coding agents under a unified protocol (abbreviated CC, OS, CX, and DS for Claude Code, OpenScience, Codex CLI, and DeepSeek Harness, respectively):
\begin{itemize}[leftmargin=1.5em,itemsep=1pt]
    \item \textbf{Claude Code} (v2.1.220) \cite{claudecode2026};
    \item \textbf{OpenScience} (v2.0.1), an open-source research-agent framework \cite{openscience2026};
    \item \textbf{Codex CLI} (v0.149.0) \cite{codexcli2026};
    \item \textbf{DeepSeek Harness} (v0.1.0rc7), an open-source agent-loop framework \cite{deepseekharness2026}.
\end{itemize}
The two open-source frameworks, OpenScience and DeepSeek Harness, are cited through their public repositories rather than archival releases, and the exact versions used here are pinned in the citations above. Claude Code and Codex CLI are released products.

All four systems use the same frozen base model, DeepSeek-V4-Flash~\cite{deepseek2026}, with thinking disabled; the scoring model is fixed as GLM-5.1~\cite{glm2026}, run in its 4-bit-weight, 8-bit-activation (W4A8) quantized configuration with thinking disabled for every run. The quantized, thinking-disabled judge is a deliberate fixed setup: it reduces inference cost and removes chain-of-thought length variance from the 0/0.5/1 item decisions, yielding a single reproducible scoring setup; a small non-quantized and thinking-enabled judge-consistency audit is left to future work (Section~\ref{sec:discussion}). Holding the base model constant is deliberate: it isolates the contribution of the agent \emph{scaffold} (tool use, state management, task decomposition, workflow) to discovery behavior, avoiding the base-model-and-scaffold confound that is present when different agents are paired with different frontier models. All systems receive identical task data, time boundaries, neutral research goals, literature-use boundaries, and output requirements, while retaining their native tool calling, state management, task decomposition, and workflows.

\paragraph{Run protocol.}
Each of the 40 tasks is run once per system, yielding 160 system--task evaluation units. Each system submits a final report, its analysis code, and its result files. The judge verifies execution results and completes the 29 per-discovery items, then performs semantic de-duplication and computes task scores. A total of 724 reported discoveries (630 after semantic de-duplication) were judged across the four systems, corresponding to $724\times29\approx2.1\times10^{4}$ item-level judgments (Section~\ref{sec:breadth}). Per-system prompts, decoding parameters, resource and tool-call limits, network permissions, and failure-handling rules were frozen internally for this study and are not part of the public package; the released repository contains the tasks, the frozen data, and the run-freezing and scoring programs, together with the evaluator assets needed to score any frozen run under the same 29-item contract. The four systems' own run artifacts and the third-party agent systems themselves are not redistributed.

\paragraph{Scoring procedure.}
For each discovery, the 29 items are scored from the actually executed analyses and artifacts; proposed-but-unexecuted tests earn no credit, and the items avoid free-form preference judgments so that scores can be regenerated automatically on arbitrarily many new instances (Section~\ref{sec:aggregation}). System means use round-half-up to two decimals, per-task and per-domain scores to one decimal, and normalized dimension rates to two decimals. A claim asserted as true in the report while the executed result or a key re-derivation clearly contradicts its direction is flagged as a deterministic error and penalized.

\subsection{Overall results: a performance plateau}
\label{sec:results}
\begin{figure}[t]
\centering
\includegraphics[width=0.68\textwidth]{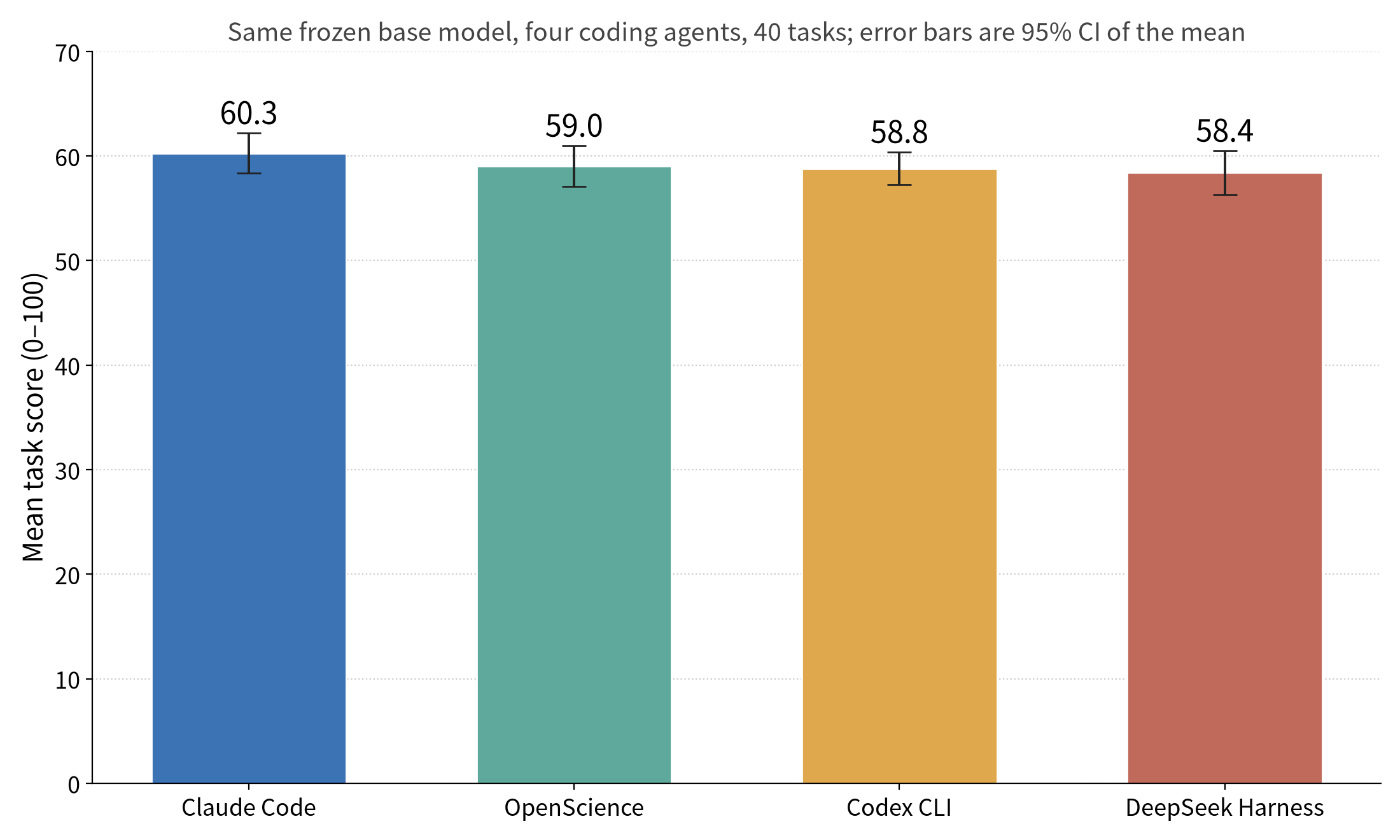}
\caption{\textbf{Mean task scores of the four coding agents (same frozen base model).} Error bars are 95\% confidence intervals of the mean (mean$\pm$1.96 SE). All four systems cluster between 58.4 and 60.3 of 100 and remain well short of credible discovery as operationalized in Section~\ref{sec:benchmark}.}
\label{fig:overall}
\end{figure}

Table~\ref{tab:overall} reports the four systems' mean task scores, 95\% confidence intervals (CIs) of the mean, ranges, between-task standard deviations (SDs), and counts of first- and second-place tasks. Figure~\ref{fig:overall} visualizes the means and their 95\% confidence intervals.

\begin{table}[t]
\centering
\small
\caption{\textbf{Overall results on \tb{} (40 tasks, equal-weighted).} Full score is 100. The 95\% interval is the confidence interval of the mean (mean$\pm$1.96 standard errors across the 40 tasks); it is not an estimate of run-to-run variance within a task. The \#1 and \#1--2 counts credit every system tied for that rank.}
\label{tab:overall}
\begin{tabular}{lcccccc}
\toprule
System & Mean & 95\% CI of mean & Min--Max & Between-task SD & \#1 & \#1--2 \\
\midrule
Claude Code (v2.1.220) & \textbf{60.27} & 58.36--62.18 & 44.3--72.9 & 6.16 & 13 & 26 \\
OpenScience (v2.0.1) & 59.03 & 57.08--60.97 & 36.2--67.4 & 6.28 & 9 & 21 \\
Codex CLI (v0.149.0) & 58.81 & 57.25--60.36 & 46.8--70.8 & 5.03 & 10 & 18 \\
DeepSeek Harness (v0.1.0rc7) & 58.40 & 56.31--60.49 & 43.8--72.2 & 6.75 & 8 & 15 \\
\bottomrule
\end{tabular}
\end{table}

All four coding agents cluster in a narrow band of 58.4--60.3 out of 100 (a spread of only 1.87 points), and their mean confidence intervals overlap. Paired two-sided Wilcoxon signed-rank tests on the 40 per-task differences find no pair that separates at the 0.05 level (smallest $p=0.118$, Claude Code versus DeepSeek Harness; all other $p\geq0.14$; a Holm correction across the six pairwise comparisons leaves these conclusions unchanged), so under a single frozen run the nominal ordering is not statistically reliable and no system is demonstrably superior at the task level. A paired equivalence analysis (TOST)~\cite{schuirmann1987} leads to the same conclusion: using a prespecified practical-equivalence margin of $\pm4$ points (four percent of the full scale and an order of magnitude below the 38--40-point strong--weak dimension gap; Section~\ref{sec:breadth}), all six pairs are equivalent (the widest 90\% confidence interval of a mean paired difference is $[-0.06,+3.79]$ for Claude Code versus DeepSeek Harness, the other five pairs are narrower, and three of the six pairs also lie within a tighter $\pm3$-point margin), while tighter run-level equivalence must await the multi-seed protocol in Section~\ref{sec:limitations}. The outcome is a performance plateau rather than a ranking.

The plateau also lies well below credible discovery. Averaged over all reported discoveries (the per-discovery rates of Table~\ref{tab:sixdim}), the systems attain only about 38--40 of the 75 points carried by the current-evidence component (evidence auditability, robustness, and control testing; Section~\ref{sec:benchmark}), or 51--53\% of that component, so a mean in the high 50s signals that the leading discovery's evidence remains incomplete along several audited dimensions.

Finally, the tight means do not indicate a benchmark that fails to discriminate. The full $40\times4$ per-task scores (Table~\ref{tab:full}) span 36.2--72.9, and on individual tasks the four systems differ by as much as 25.5 points (Section~\ref{sec:difficulty}); each system also tops a comparable number of tasks (Table~\ref{tab:overall}, 8--13 first-place finishes). The narrow means instead arise from compensating wins and losses across tasks, not from uniform performance or from any single dominant system.

\subsection{The discovery--execution gap}
\label{sec:gap}
The narrow range of overall means conceals a sharply structured profile: the systems generate claims at a comparable rate, but the evidence supporting those claims, and the discriminating tests required to establish them, is weak. We develop this diagnosis in two steps, first contrasting discovery breadth with evidence depth and then decomposing performance along the six quality dimensions along which discovery is separated from execution.

\paragraph{Breadth versus evidence depth.}
\label{sec:breadth}
Table~\ref{tab:breadth} separates discovery breadth from evidence depth. All four systems report a similar number of discoveries per task after semantic de-duplication ($\approx$3.65--4.30 per task), with repetition rates of 9.9--15.1\%, indicating that reports are largely non-redundant and the yield bonus is not driven by restating the same claim. Yet their two highest-quality discoveries attain only 43.3--44.6 of a possible 80 points in evidence quality. In other words, current coding agents generate comparable numbers of candidate claims, yet the claims themselves are not supported by mature evidence: their weighted top-two evidence score reaches only approximately 55\% of the available 80 points. Moreover, top-two evidence quality spans only 1.25 points across systems (43.33--44.58 of 80). Decomposing the 1.87-point spread between the highest- and lowest-scoring system shows that 1.25 points already reside in this evidence-quality body and only 0.62 points in the combined yield and reference-coverage bonuses, with the evidence-quality body alone reproducing the overall ordering; overall differences are thus small on every component, and no component supports a stable system ranking.

\begin{table}[t]
\centering
\small
\caption{\textbf{Discovery breadth and top-two evidence quality.} Reported discoveries are all claims in the final report; de-duplicated discoveries are semantically merged groups; the top-two contribution is $0.60S_1+0.20S_2$ averaged over 40 tasks (maximum 80); the repeat rate is $1-\mathrm{dedup./reported}$ within each system.}
\label{tab:breadth}
\begin{tabular}{lccccc}
\toprule
System & Reported & Dedup. & Per task & Repeat rate & Top-two quality (max 80) \\
\midrule
Claude Code & 191 & 172 & 4.30 & 9.9\% & \textbf{44.58} \\
OpenScience & 172 & 149 & 3.73 & 13.4\% & 44.17 \\
Codex CLI & 172 & 146 & 3.65 & 15.1\% & 44.11 \\
DeepSeek Harness & 189 & 163 & 4.08 & 13.8\% & 43.33 \\
\bottomrule
\end{tabular}
\end{table}

\begin{figure}[t]
\centering
\includegraphics[width=0.95\textwidth]{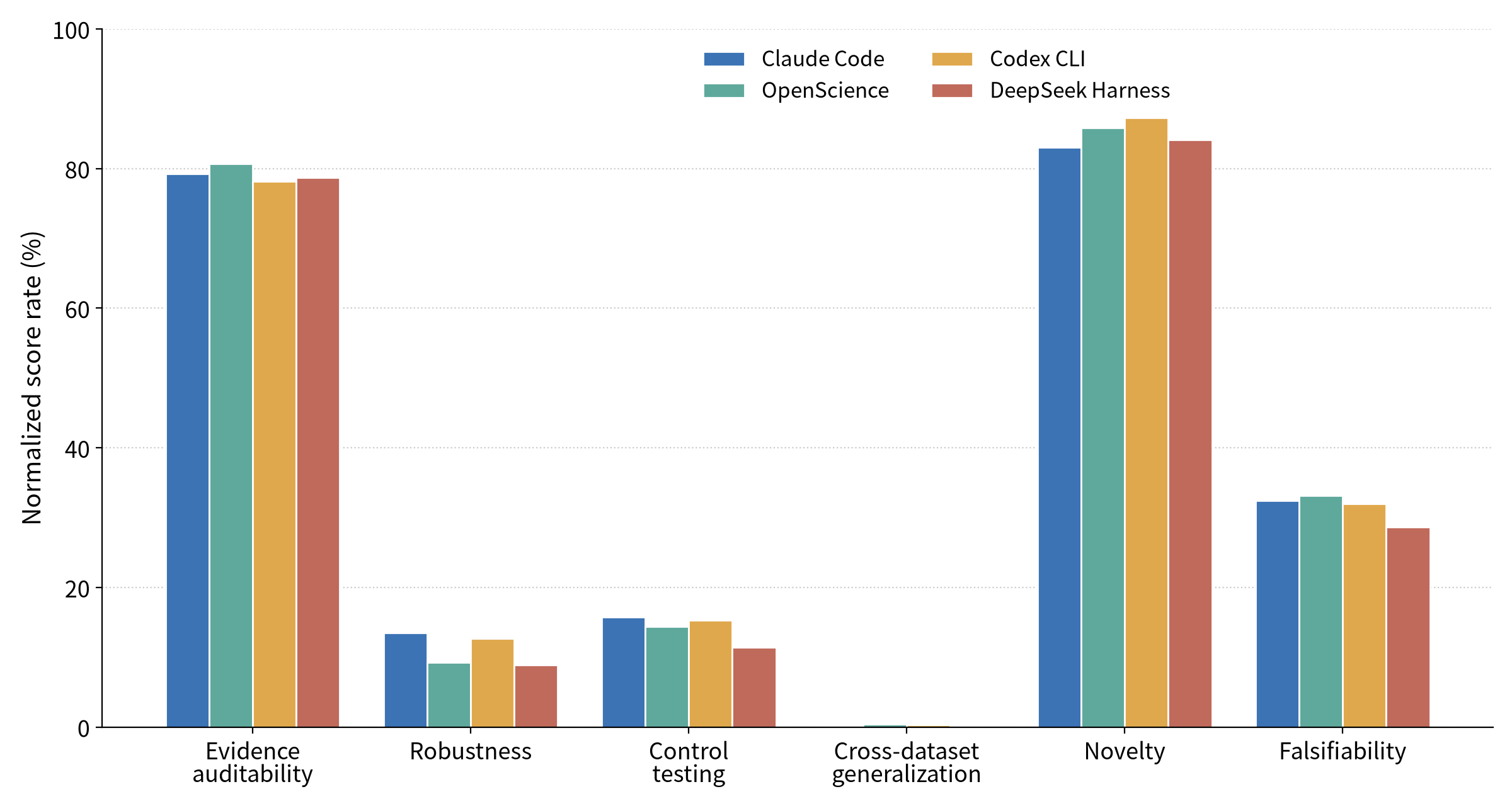}
\caption{\textbf{Six-dimension normalized score rates across the four coding agents.} Evidence auditability and novelty are comparatively strong; control testing, robustness, falsifiability, and cross-dataset generalization are weak. The discovery--execution gap is precisely the gap between these two groups.}
\label{fig:sixdim}
\end{figure}

\paragraph{Six-dimension diagnostic.}
\label{sec:sixdim}
Table~\ref{tab:sixdim} and Figure~\ref{fig:sixdim} report normalized score rates for the six dimensions, computed as the sum of achieved points over all reported discoveries divided by (number of discoveries $\times$ dimension maximum), aggregated by reported discovery rather than by task. Grouping the six dimensions by their observed levels, a cut distinct from the conceptual 75/20/5 blocks of Section~\ref{sec:benchmark}, the two high-scoring dimensions (evidence auditability and novelty, which carry $45+10=55$ of the 100 points) attain about 44, whereas the four low-scoring dimensions (control testing, robustness, falsifiability, and cross-dataset generalization, which carry $15+15+5+10=45$ points) attain only about 4.5--6; this is the 38--40-point strong--weak gap cited in Section~\ref{sec:results}.

\begin{table}[t]
\centering
\small
\caption{\textbf{Six-dimension normalized score rates (\%, over all reported discoveries).} The gap between execution and discovery concentrates in control testing, robustness, falsifiability, and cross-dataset generalization.}
\label{tab:sixdim}
\footnotesize
\setlength{\tabcolsep}{4pt}
\begin{tabular}{>{\raggedright\arraybackslash}p{2.7cm}ccccc>{\raggedright\arraybackslash}p{2.4cm}}
\toprule
Dimension & Claude Code & OpenScience & Codex CLI & DeepSeek & Note \\
\midrule
Evidence\newline auditability & 79.25 & 80.68 & 78.19 & 78.74 & data-to-claim chain \\
Robustness & 13.51 & 9.24 & 12.67 & 8.89 & near-universally low \\
Control\newline testing & 15.78 & 14.40 & 15.29 & 11.45 & largest diagnostic gap \\
Cross-dataset\newline generalization & 0.10 & 0.41 & 0.29 & 0.00 & $\approx$0 under single-phase tasks \\
Novelty & 83.04 & 85.87 & 87.27 & 84.13 & mostly self-formed from data \\
Falsifiability & 32.46 & 33.14 & 31.98 & 28.62 & update rules rarely executable \\
\bottomrule
\end{tabular}
\end{table}

The pattern is pronounced and consistent across all four systems:
\begin{itemize}[leftmargin=1.5em,itemsep=1pt]
    \item \textbf{Evidence auditability is comparatively strong (78--81\%).} Coding agents can execute an analysis, keep its artifacts, and produce a report whose data--execution--result--claim chain is largely traceable and verifiable.
    \item \textbf{Novelty is comparatively strong (83--87\%).} Since no endpoint is given, the systems' claims are mostly formed from the data themselves rather than copied from external answers, consistent with the blind protocol functioning as designed.
    \item \textbf{Control testing is weak (11--16\%).} Systems rarely run the discriminating tests that would establish that their explanation is specific rather than generic: simple baselines, no-signal controls, competing explanations, cross-method corroboration, and confound/artifact checks.
    \item \textbf{Robustness is weak (9--14\%),} with the few earned points coming mainly from stating applicability boundaries rather than quantifying uncertainty; explicit uncertainty quantification is rarer still.
    \item \textbf{Falsifiability is low (29--33\%),} meaning systems rarely specify what future observation would maintain, narrow, or overturn their conclusions in a runnable form.
    \item \textbf{Cross-dataset generalization is near zero for three distinct reasons.} The dimension is not structurally forced to zero: its means are 0.10/0.41/0.29/0.00 rather than identically zero, and the non-zero upper bound (OpenScience 0.41) shows the rubric awards points when independent validation is actually performed. (A)~\emph{Benchmark-side floor:} some tasks ship a single data block with no natural second source, time window, or population on which to hold out. (B)~\emph{A genuine agent-side discipline gap:} on tasks whose released data do contain repeats, variation, or controls and could be split, agents still rarely construct a held-out set, freeze criteria beforehand, or perform out-of-sample validation, the most consequential component. (C)~\emph{Single-phase operational limit:} the pre-specified-criteria and discovery--validation-separation sub-items require temporal guarantees that one continuous session cannot evidence. We retain the dimension and read its near-zero level mainly as the rarity of self-imposed out-of-sample validation; a task-level feasibility classification and a two-phase protocol separating (A)--(C) are left to future work.
\end{itemize}

This constitutes the central finding of this study. On a benchmark that requires discovery rather than reproduction, mainstream coding agents fall short not in code generation or output traceability, capabilities at which they are comparatively proficient, but in the scientific acts that distinguish discovery from execution: testing whether a simpler explanation suffices, verifying whether an analysis would still report a phenomenon on negative-control data containing no signal, comparing competing mechanisms, and specifying the observations that would falsify their claims.

\subsection{Domain and task-level structure}
\begin{figure}[t]
\centering
\includegraphics[width=0.72\textwidth]{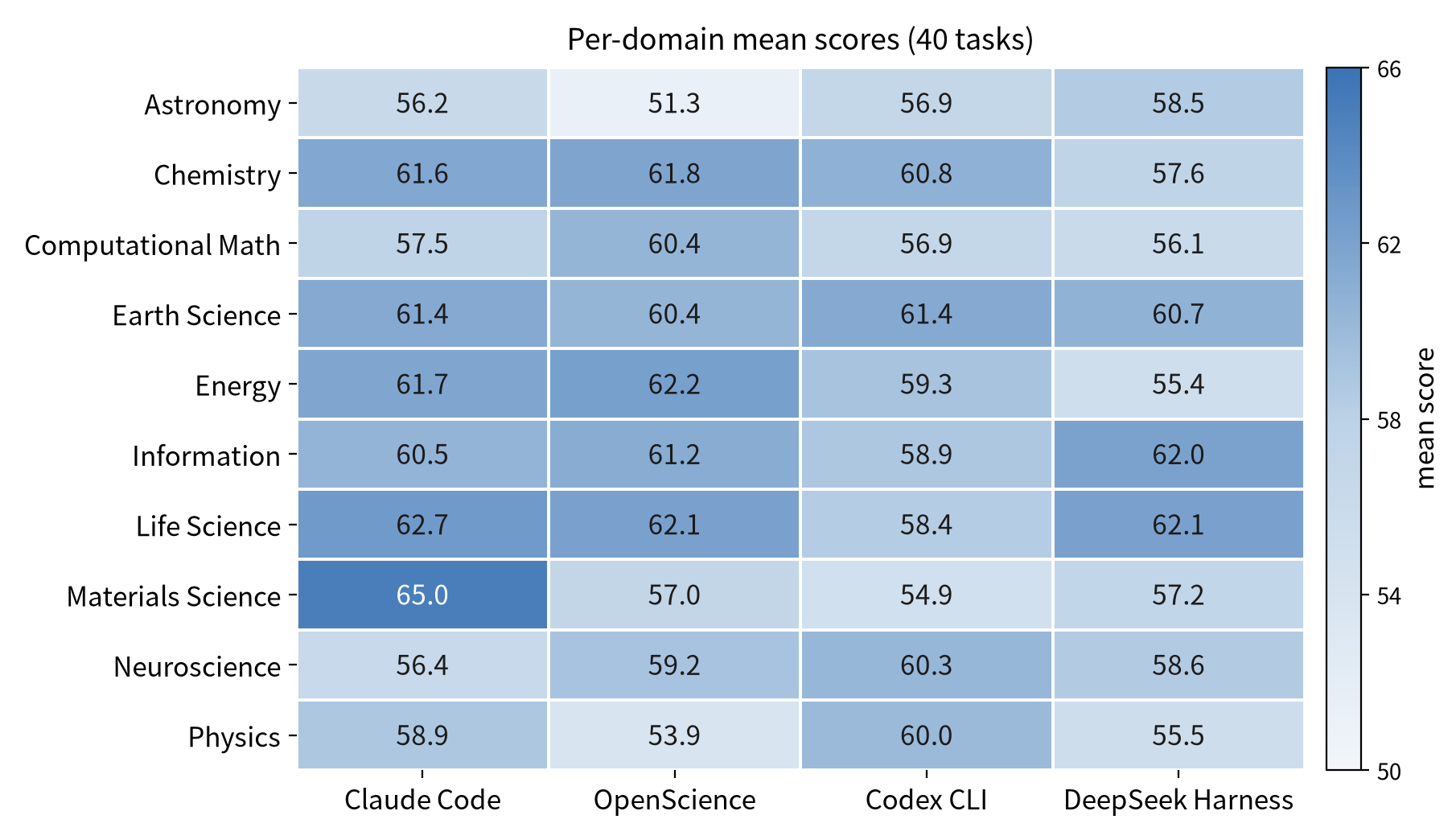}
\caption{\textbf{Per-domain mean scores (10 domains $\times$ 4 coding agents).} Darker cells indicate higher mean scores. No system leads all domains.}
\label{fig:domain}
\end{figure}

Figure~\ref{fig:domain} reports the per-domain mean scores. The identity of the top-scoring system rotates across domains, with each system highest in two or three fields, but this rotation should not be overinterpreted: the top system's margin over the runner-up is at or below 1.1 points in seven of the ten domains (six strictly below; median 0.90), only a small fraction of the between-task standard deviation (5.03--6.75), and at three to five tasks per domain such differences lie within task-level noise. Only Materials Science, where Claude Code leads by 7.72 points, separates a single system clearly. We therefore make no claim of stable domain specialization. Together with the tight overall clustering and the non-significant pairwise tests of Section~\ref{sec:results}, this rotation of nominal leaders on noise-level margins reinforces the plateau result: under one frozen base model, no system holds a general discovery advantage. This conclusion is also robust to aggregation: an equal-weighted average of the ten domain means (rather than the 40 tasks) gives 60.2/59.0/58.8/58.4 for Claude Code/OpenScience/Codex CLI/DeepSeek and preserves the same ordering, so the plateau is not an artifact of unequal domain sizes.

\begin{figure}[ht]
\centering
\includegraphics[width=0.98\textwidth]{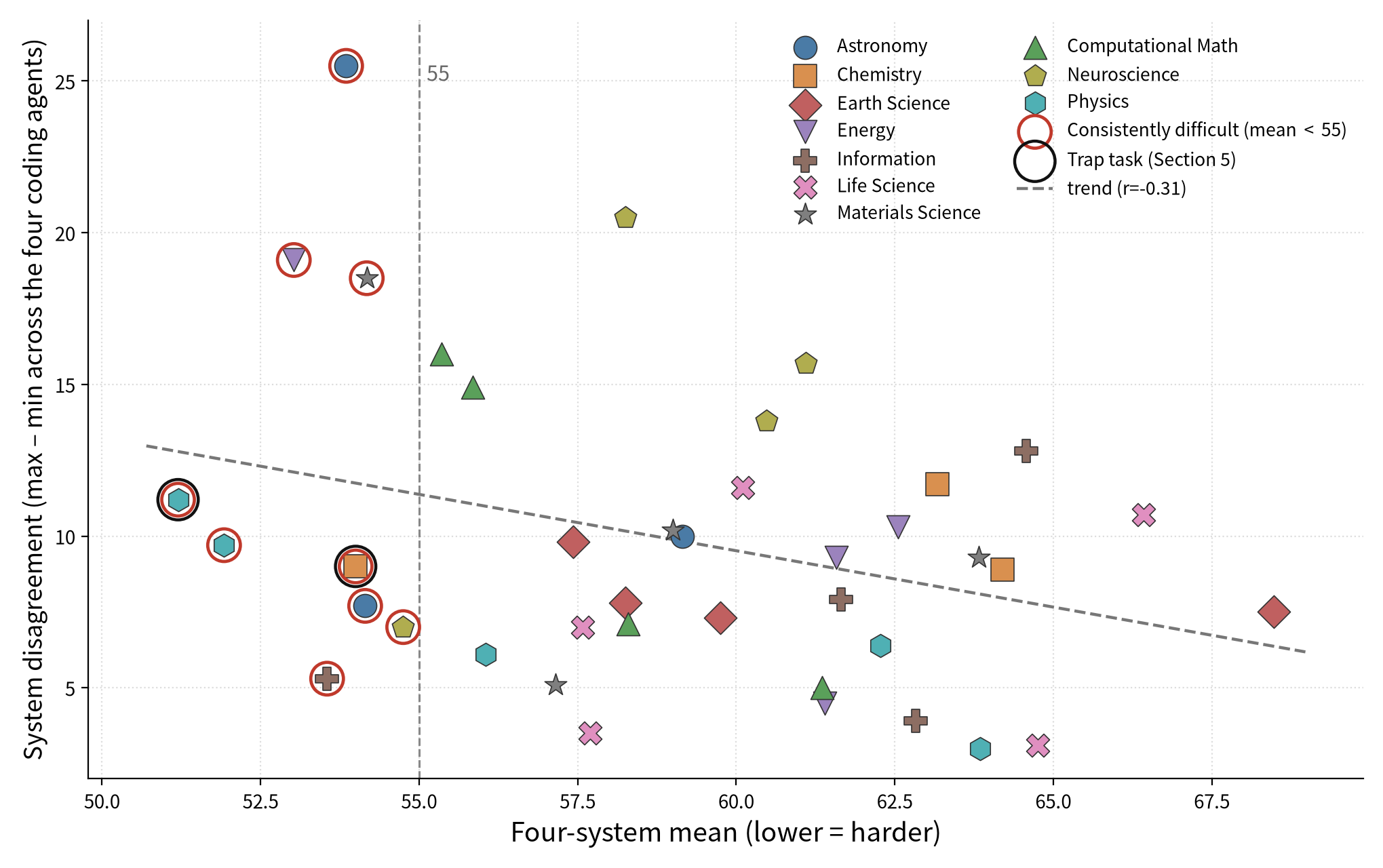}
\caption{\textbf{Task difficulty versus system disagreement} across the 40 tasks. The horizontal axis is the four-system mean (lower = harder); the vertical axis is the max--min gap among the four systems. Harder tasks (four-system mean below 55, i.e., left of the vertical dashed line) tend to produce the largest system disagreement, which is where discovery-oriented scoring has the most discriminative power. Red outlines mark the consistently difficult tasks; black outlines mark the ``trap'' tasks whose apparent signal is a numerical or modeling artifact. The exact per-task scores are given in Appendix~\ref{app:pertask}.}
\label{fig:scatter}
\end{figure}

\paragraph{Task-level difficulty and discrimination.}
\label{sec:difficulty}
Figure~\ref{fig:scatter} plots each task by difficulty against system disagreement, and the full $40\times4$ per-task score matrix is given in Appendix~\ref{app:pertask}. Although the overall means occupy a narrow range, individual tasks reveal substantial between-system differences and a consistent difficulty structure. To characterize task difficulty independently of any single system, we define a task's baseline mean as the average of the four systems' scores on it. Nine tasks have a baseline mean below 55. In ascending order of difficulty they are Physics\_04 (51.2; temporal-delay transition with a synthetic-to-real-network generalization trap), Physics\_01 (51.9; dynamic nuclear polarization echoes), Energy\_03 (53.0; interface response boundaries), Information\_03 (53.6; signal-surface invariants), Astronomy\_01 (53.9; radio-burst drift, on which OpenScience attains 36.2), Chemistry\_02 (54.0; photoionization delay with interpolation-artifact traps), Astronomy\_03 and Material\_02 (54.2; stellar spectral boundary and temperature-response boundaries, respectively), and Neuro\_01 (54.8; self-organization prediction); Math\_01 (55.4; governing-equation discovery, on which Claude Code attains 44.3) lies just above the threshold. These are precisely the tasks that reward discriminating scientific acts (distinguishing a real signal from a numerical artifact, delimiting an extrapolation boundary, or discovering the governing form) rather than executing a given pipeline.

Conversely, the benchmark does not saturate at the high end: even its highest-scoring tasks remain well below a mature discovery score, and per-task spreads reach 25.5 points (Astronomy\_01: 36.2--61.7; Neuro\_04: 44.9--65.4; Material\_02: 46.8--65.3; Math\_01: 44.3--60.3). The tight overall means therefore reflect compensating wins and losses across systems, not a benchmark that cannot separate systems. As Figure~\ref{fig:scatter} shows, a task's mean score and its cross-system disagreement are weakly and negatively correlated (Pearson $r=-0.31$, $p=0.049$; Spearman $\rho=-0.23$, $p=0.15$): lower-scoring, i.e., harder, tasks tend to show somewhat wider divergence. This is a descriptive observation only: Spearman is not significant, and the mean and the range share the same four scores, which induces some mathematical coupling. Discovery-oriented scoring nonetheless separates systems most clearly on harder tasks, precisely where discriminating scientific acts are required.

\section{Diagnostic Case Studies}
\label{sec:errors}

The six-dimension profile of Section~\ref{sec:gap} locates the deficit quantitatively, while Section~\ref{sec:difficulty} and Figure~\ref{fig:scatter} identify the tasks on which it is most pronounced. Here we examine two such ``trap'' tasks qualitatively: their data invite a superficially compelling but unwarranted reading under naive execution, and the correct scientific act is a discriminating test.

\paragraph{Chemistry\_02 (photoionization delay).}
The delayed-response curves contain extreme spikes that, on superficial inspection, resemble a new physical resonance. Whether the spikes are physical or a numerical artifact of high-order interpolation can only be decided by comparing interpolation methods and checking stability under node perturbation, the very control-testing acts on which all four coding agents score near zero in aggregate. The systems score 50.9--59.9 on this task, below or near their means, separating those that verify their methods from those that report the surface pattern.

\paragraph{Physics\_04 (temporal-delay transition).}
The task asks whether an apparent critical slowdown generalizes from synthetic models to real networks. The systems score 45.6--56.8, among the lowest-scoring regions of the benchmark, consistent with reporting the visually salient synthetic trend without testing whether it survives on real data; the task measures whether a system can delimit its own extrapolation.

These tasks are not adversarial trick items but ordinary scientific situations in which the correct act is a discriminating test. Their presence is what lets \tb{} measure discovery rather than execution.

\section{Discussion}
\label{sec:discussion}

\subsection{Scope: what \tb{} measures and does not measure}
\tb{} measures data-driven discovery ability and discovery quality in a \emph{retrospective, blind} setting: systems must re-form and re-test scientific claims from frozen data without knowing the source conclusions or the standard analysis path. We do not claim that a high score constitutes a real-world first discovery; retrospective data-driven claims still require new data, new experiments, and independent teams for ultimate confirmation. Three further boundaries deserve emphasis:
\begin{itemize}[leftmargin=1.5em,itemsep=1pt]
    \item \textbf{Protocol-level blindness.} The protocol proves that source conclusions and evaluation assets never entered the task workspace, but it cannot prove that a base model never encountered related papers during pretraining. We therefore claim protocol-level blindness, not absolute contamination freedom. The fixed literature cutoff $T_0$ and the novelty-scoring rule (comparing only pre-$T_0$ literature) mitigate, but cannot fully eliminate, this concern.
    \item \textbf{Static single-phase tasks.} The current protocol provides limited second-phase data, so the cross-dataset-generalization dimension is under-measured (near-zero for all systems). Two-phase tasks that release validation data only after a discovery and its pre-specified criteria are recorded are a priority for future work.
    \item \textbf{Single frozen run per system--task.} Each system--task cell contains one frozen run. Open-ended discovery is stochastic (re-running the same system on the same data can select different claims to pursue and hence a different set of discoveries), so it is useful to separate two variance layers. \emph{Scoring variance} (how reproducibly a fixed set of reported discoveries is judged) is constrained by design: the judge itself is frozen, aggregation is deterministic, and each item is scored against the executed analyses and artifacts rather than free-form text. \emph{Selection variance} (which claims the system forms and how deeply it tests them) is a property of the agent; it is not estimated under a single run, and we treat it as a measurement target rather than as pure noise; what the single-run design can and cannot support, and the planned multi-seed protocol, are discussed in Section~\ref{sec:limitations}.
\end{itemize}

\subsection{Toward recursive research-improvement loops}
\tb's automated, artifact-grounded score is designed to function as the feedback channel in an automated research-improvement loop: an agent runs on a task, receives dimension-level scores and item-level diagnostics without any human in the per-instance path, revises its research procedure, and re-runs at scale. Because a single-run total is a noisy per-instance signal, the loop should consume the more stable aggregates: the mean over $k$ repeated runs of the same system--task, or the dimension-level diagnostics, which pool over all of a system's reported discoveries and are therefore markedly more stable than any single-run total (Section~\ref{sec:limitations}). We view \tb{} not as a demonstration of recursive self-improvement but as one necessary, automatable building block for it: a discovery-oriented evaluation whose feedback can be regenerated automatically on every new instance.

\subsection{Limitations and future work}
\label{sec:limitations}
Several scope boundaries delimit this release and define the immediate next steps. All four coding agents share one frozen, mid-capability base model and are evaluated in a single run, so the study isolates scaffold effects under that model and supports the broad execution-versus-discovery profile, whose roughly 38--40-point gap between the strong and weak dimensions far exceeds any plausible run-level noise (Section~\ref{sec:breadth}), but not exact pairwise ordering, tight-margin equivalence, or whether identical gaps persist with frontier models; multi-seed and multi-model studies are the natural next step, and the planned protocol re-runs each system--task with fixed seeds where the configuration permits and reports the mean over $k$ repeated runs alongside the per-run spread. Tasks are computational (``dry-lab'') and drawn from sources with open, reproducible data, so transfer to wet-lab research and to less curated corpora remains to be established. Automated scoring by a single frozen LLM judge~\cite{zheng2023} is a deliberate scalability requirement for self-improving research loops rather than a claim that automated judgment equals expert judgment; a human-expert calibration and a second-judge sensitivity analysis are planned, after which routine evaluation remains fully automatic. Future versions will also broaden modalities and pair open release with a rolling held-out set so that the blind condition is preserved as agents evolve.

\section{Conclusion}
\label{sec:conclusion}

We presented \tb, an exploratory benchmark configured for open-ended scientific discovery rather than reproduction. Where reproduction-oriented benchmarks organize a task, its data, and its rubric around a hidden target study, \tb{} gives an agent only a neutral scientific objective and frozen data, hides all source conclusions and evaluation assets, and scores the evidentiary maturity of the agent's own claims along six dimensions and 29 artifact-grounded items. Because scoring is automated and its aggregation deterministic, the same evaluation can be repeated, without per-instance human grading, on every new agent version, a property we regard as essential if evaluation is to keep pace with self-improving research systems. Null results, counterexamples, method artifacts, and scope boundaries are credited on the same footing as positive regularities, so the instrument rewards scientific judgment rather than recovery of a privileged endpoint.

A controlled study of four mainstream coding agents on a single frozen base model offers a preliminary validation of the instrument and a clear diagnostic. Total scores form a narrow plateau (58.4--60.3 of 100) with no statistically reliable pairwise separation, and the systems remain well short of credible discovery even as they conduct and document analyses competently: the deficit is concentrated in the discriminating acts that separate discovery from execution (controls, robustness, falsifiability, and cross-dataset validation), not in coding or traceability. The bottleneck for current AI-scientist systems is therefore scientific judgment rather than execution. We offer this four-system comparison as an initial snapshot rather than a final ranking; multi-seed replication, additional base models, human-expert anchors, and judge-reliability audits are the natural next steps. The task data and scoring program are openly released, and we hope \tb{} provides a reproducible, automatable instrument for measuring and closing the gap between executing analyses and making discoveries as autonomous research agents evolve.

\appendix
\clearpage
\section{Authors}
\label{app:authors}
\noindent\textbf{Core Authors.} Zhibo Yang\textsuperscript{2}, Chen Zhang\textsuperscript{2}, Yuewei Zhang\textsuperscript{3}, and Hao (Henry) Wang\textsuperscript{1}.

\vspace{0.6em}
\noindent\textbf{Main Affiliations.}\\
\textsuperscript{1}~Computer Network Information Center, Chinese Academy of Sciences\\
\textsuperscript{2}~Jingdong Group\\
\textsuperscript{3}~Alibaba Cloud
\clearpage

\section{The 29 evaluation items}
\label{app:items}
Each item is judged as fully satisfied ($c_j=1$), partially satisfied ($c_j=0.5$), or unsatisfied ($c_j=0$). Unless otherwise stated, \emph{fully satisfied} requires all frozen essential elements or steps to be completed; \emph{partially satisfied} requires that no key conflict affecting the conclusion's direction exists and that the frozen weighted coverage reaches 50\% but not 100\%; below 50\% or with any direction-critical conflict, the item is \emph{unsatisfied}. Weights shown are the item's maximum contribution to the 100-point discovery quality score.

\subsection*{B.1 Evidence auditability (45 points)}
\begin{center}
\small
\begin{tabular}{p{3.4cm}cp{6.3cm}}
\toprule
Item & Weight & Evaluation focus \\
\midrule
Data consistency & 4 & Data source, version, and scope match the task; conclusion-critical data are correct and traceable. \\
Object, scope, unit consistency & 4 & Study object, analysis scope, units, and independent-observation units are correctly defined. \\
Analysis-execution validity & 6 & Key analyses actually ran successfully; report matches actual results. \\
Result traceability & 4 & Every claim and key number maps to a concrete analysis result. \\
Core-judgment reproducibility & 10 & Key quantities that determine conclusion direction/category/boundary reproduce the reported conclusion from the executed analysis. \\
Key-number reproducibility & 5 & Effect sizes, sample sizes, and supporting numbers reproduce within tolerance from the executed analysis. \\
Claim--evidence consistency & 8 & Executed results support the claimed object, scope, relation type, direction, and strength. \\
Synthesis of all results & 4 & Final conclusion faithfully absorbs successes, failures, and null results of later tests. \\
\bottomrule
\end{tabular}
\end{center}

\subsection*{B.2 Robustness (15 points)}
\begin{center}
\small
\begin{tabular}{p{3.4cm}cp{6.3cm}}
\toprule
Item & Weight & Evaluation focus \\
\midrule
Uncertainty quantification & 3 & Main random, measurement, or model uncertainty estimated with applicable methods. \\
Stochasticity sensitivity & 3 & Complete randomized reruns under specified seeds, retained. \\
Sample sensitivity & 3 & Batch, outlier, held-out, or time-split perturbations, retained. \\
Analysis-choice sensitivity & 3 & Method, parameter, or threshold variations, retained. \\
Applicability boundary & 3 & Stated regime of validity, observed failure, and untested scope consistent with perturbations. \\
\bottomrule
\end{tabular}
\end{center}

\subsection*{B.3 Control testing (15 points)}
\begin{center}
\small
\begin{tabular}{p{3.4cm}cp{6.3cm}}
\toprule
Item & Weight & Evaluation focus \\
\midrule
Simple-baseline comparison & 3 & Target explanation actually compared with constant/null/linear/exponential or other simple models. \\
No-signal control & 3 & Blank, random-label, no-treatment, or unrelated-object control checks the method's false-positive rate. \\
Competing-explanation test & 4 & A principal competing mechanism is actually tested with discriminating predictions. \\
Confound / artifact check & 2 & Grouping, batch, sampling, measurement, preprocessing, or fitting-scale confounds checked. \\
Cross-method corroboration & 3 & A second analysis with different dominant error sources corroborates the conclusion. \\
\bottomrule
\end{tabular}
\end{center}

\subsection*{B.4 Cross-dataset generalization (10 points)}
\begin{center}
\small
\begin{tabular}{p{3.4cm}cp{6.3cm}}
\toprule
Item & Weight & Evaluation focus \\
\midrule
Validation-data independence & 2 & Validation data did not participate in hypothesis, feature, threshold, or model selection. \\
Cross-dataset execution & 3 & Pre-registered analysis actually executed on new data/time/sources/objects. \\
Pre-specified criteria & 3 & Metrics, thresholds, and conclusion-update rules fixed before viewing new-data results. \\
Discovery--validation separation & 2 & No direction-critical observations or answer-derived labels fed back into the discovery route. \\
\bottomrule
\end{tabular}
\end{center}

\subsection*{B.5 Novelty (10 points)}
\begin{center}
\small
\begin{tabular}{p{3.4cm}cp{6.3cm}}
\toprule
Item & Weight & Evaluation focus \\
\midrule
Formation independence & 4 & Core claim formed from task data; external material used only for background and comparison. \\
Knowledge increment & 6 & Against the frozen pre-$T_0$ literature, the claim adds an object, relation, result, or boundary. \\
\bottomrule
\end{tabular}
\end{center}

\subsection*{B.6 Falsifiability (5 points)}
\begin{center}
\small
\begin{tabular}{p{3.4cm}cp{6.3cm}}
\toprule
Item & Weight & Evaluation focus \\
\midrule
Future evidence source & 1 & Specifies new data, observation, or experiment that can further test the claim. \\
Discriminating measurement & 1 & A measurement that makes principal competing explanations produce distinguishable predictions. \\
Conclusion-update condition & 1 & States what observation maintains, strengthens, narrows, or overturns the conclusion. \\
Implementation completeness & 1 & Required data, measurement, compute, permissions, time, and budget are in place. \\
Executable decision rule & 1 & The conclusion-update rule is fixed in advance, runnable, and covers the outcome space. \\
\bottomrule
\end{tabular}
\end{center}

\section{Complete per-task score matrix (40 tasks $\times$ four coding agents)}
\label{app:pertask}
Table~\ref{tab:full} reports the 160 frozen task scores (40 tasks $\times$ 4 systems). Bold marks the per-task maximum.

\begingroup
\scriptsize
\setlength{\tabcolsep}{4pt}
\renewcommand{\arraystretch}{1.05}
\begin{longtable}{p{7.4cm}cccc}
\caption{\textbf{Full per-task scores (40 tasks $\times$ 4 systems).} Columns: CC=Claude Code, OS=OpenScience, CX=Codex CLI, DS=DeepSeek Harness. Bold marks the per-task maximum among the four systems.}\label{tab:full}\\
\toprule
Task & CC & OS & CX & DS \\
\midrule
\endfirsthead
\caption[]{\textbf{Full per-task scores (continued).}}\\
\toprule
Task & CC & OS & CX & DS \\
\midrule
\endhead
\midrule
\multicolumn{5}{r}{\itshape Continued on the next page}\\
\endfoot
\bottomrule
\endlastfoot
Astronomy 01 radio burst drift & 56.5 & 36.2 & 61.0 & \textbf{61.7} \\
Astronomy 02 gravity spectral structure & \textbf{63.0} & 61.8 & 53.0 & 58.8 \\
Astronomy 03 stellar spectral boundary & 49.0 & 55.9 & \textbf{56.7} & 55.0 \\
Chemistry 01 bond enthalpy domain & 64.1 & \textbf{67.4} & 58.5 & 66.8 \\
Chemistry 02 photoionization delay & 54.1 & 51.1 & \textbf{59.9} & 50.9 \\
Chemistry 03 catalysis response landscape & 66.5 & \textbf{66.9} & 64.1 & 55.2 \\
EarthSci 01 glacial fluvial erosion & \textbf{71.8} & 64.3 & 70.8 & 67.0 \\
EarthSci 02 ice particle composition & 56.6 & \textbf{62.8} & 55.0 & 58.6 \\
EarthSci 03 ice core event structure & \textbf{61.4} & 51.6 & 57.6 & 59.1 \\
EarthSci 04 ocean thermal coupling & 55.7 & \textbf{63.0} & 62.2 & 58.1 \\
Energy 01 photocatalytic tradeoffs & 59.0 & 59.7 & \textbf{68.3} & 59.3 \\
Energy 02 flow battery op.\ boundary & \textbf{64.7} & 60.2 & 60.3 & 60.4 \\
Energy 03 interface response bounds & 54.9 & \textbf{62.9} & 50.5 & 43.8 \\
Energy 04 biohybrid op.\ tradeoffs & \textbf{68.2} & 65.8 & 58.3 & 57.9 \\
Information 01 dynamics general. & 61.7 & 63.4 & 56.8 & \textbf{64.7} \\
Information 02 network path bounds & 61.9 & 64.1 & \textbf{64.6} & 60.7 \\
Information 03 signal surface invar. & 53.4 & \textbf{55.6} & 54.9 & 50.3 \\
Information 04 connectome assort. & 64.8 & 61.9 & 59.4 & \textbf{72.2} \\
Life 01 dual sensor fusion & \textbf{72.9} & 64.8 & 62.2 & 65.8 \\
Life 02 contextual immune response & \textbf{66.6} & 64.7 & 63.5 & 64.2 \\
Life 03 nutrient response kinetics & 56.2 & 58.4 & 56.5 & \textbf{59.7} \\
Life 04 environmental transfer bnd. & 59.6 & \textbf{66.8} & 55.2 & 58.8 \\
Life 05 developmental interaction & 58.1 & 55.6 & 54.8 & \textbf{61.8} \\
Material 01 binary superlattice & \textbf{60.1} & 58.1 & 55.4 & 55.0 \\
Material 02 temperature response & \textbf{65.3} & 54.6 & 46.8 & 50.0 \\
Material 03 membrane permeation & \textbf{70.1} & 61.3 & 63.1 & 60.8 \\
Material 04 nir emission structure & \textbf{64.4} & 54.2 & 54.2 & 63.2 \\
Math 01 governing equation disc. & 44.3 & \textbf{60.3} & 59.6 & 57.2 \\
Math 02 compositional invariants & \textbf{63.6} & 61.4 & 61.8 & 58.6 \\
Math 03 ode parameter geometry & \textbf{63.1} & 56.0 & 57.0 & 57.1 \\
Math 04 anomalous diffusion cps & 59.0 & \textbf{63.9} & 49.0 & 51.5 \\
Neuro 01 self organization pred. & 51.1 & 56.1 & \textbf{58.1} & 53.7 \\
Neuro 02 calcium response regions & 58.0 & 62.9 & 53.9 & \textbf{69.6} \\
Neuro 03 spatial code stability & 52.3 & 59.6 & 63.9 & \textbf{66.1} \\
Neuro 04 temporal geometry & 64.3 & 58.4 & \textbf{65.4} & 44.9 \\
Physics 01 dynamic nuclear pol. & 55.0 & 48.2 & \textbf{57.1} & 47.4 \\
Physics 02 multiorbital transition & 57.4 & 52.6 & \textbf{58.7} & 55.5 \\
Physics 03 controlled dyn.\ forecast & 63.8 & 62.5 & \textbf{65.5} & 63.6 \\
Physics 04 temporal delay transition & 55.3 & 47.1 & \textbf{56.8} & 45.6 \\
Physics 05 variational circuit sim. & 62.9 & 59.0 & 61.8 & \textbf{65.4} \\
\midrule
\textbf{Mean (40 tasks)} & \textbf{60.27} & 59.03 & 58.81 & 58.40 \\
\end{longtable}
\endgroup

\end{document}